%% file: main.tex
\documentclass[11pt, a4paper, logo, copyright, singlecolumn]{deepmind}
\pdfoutput=1

\pdftrailerid{redacted}

\makeatletter
\renewcommand\bibentry[1]{\nocite{#1}{\frenchspacing\@nameuse{BR@r@#1\@extra@b@citeb}}}
\makeatother

\usepackage{dsfont}
\usepackage{dm-colors}
\usepackage{xcolor}
\usepackage{graphicx}
\usepackage{hyperref}
\usepackage{multirow}
\input{defns}

\usepackage{amsmath,amssymb}

\newcommand{\name}{Standardised Test Suite }
\newcommand{\acronym}{STS}

\usepackage[authoryear,round]{natbib} 

\title{Evaluating Multimodal Interactive Agents}
\author[1]{Interactive Agents Team}
\affil[1]{DeepMind}

\renewcommand{\today}{}

\usepackage{graphicx}

\begin{abstract}
Creating agents that can interact naturally with humans is a common goal in artificial intelligence (AI) research. However, evaluating these interactions is challenging: collecting online human-agent interactions is slow and expensive, yet faster proxy metrics often do not correlate well with interactive evaluation. In this paper, we assess the merits of these existing evaluation metrics and present a novel approach to evaluation called the \name (\acronym). The \acronym \space uses behavioural scenarios mined from real human interaction data. Agents see replayed scenario context, receive an instruction, and are then given control to complete the interaction offline. These agent continuations are recorded and sent to human annotators to mark as success or failure, and agents are ranked according to the proportion of continuations in which they succeed. The resulting \acronym \space is fast, controlled, interpretable, and representative of naturalistic interactions. Altogether, the \acronym \space consolidates much of what is desirable across many of our standard evaluation metrics, allowing us to accelerate research progress towards producing agents that can interact naturally with humans. \url{https://youtu.be/YR1TngGORGQ}
\end{abstract}

\begin{document}
\maketitle

\input{introduction}
\input{background}
\input{sts}
\input{results}
\input{additional_features}
\input{limitations}
\input{related_work}
\input{conclusion_eval}

\input{authors}
\newpage
\bibliographystyle{unsrtnat}
\bibliography{references}

\appendix
\input{appendix}
\end{document}

%% file: defns.tex
\usepackage{mathtools}
\usepackage{dsfont}
\usepackage[dvipsnames]{xcolor}
\usepackage[colorinlistoftodos]{todonotes}
\usepackage{booktabs}
\usepackage{xfrac}
\usepackage{bbm}

\usepackage{algorithm}
\usepackage{algorithmicx}

\usepackage[most]{tcolorbox}
\usepackage{xparse}
\usepackage{lipsum}
\usepackage{changepage}
\usepackage{enumitem}

\newcommand{\squishlist}{
   \begin{list}{$\bullet$}
    { \setlength{\itemsep}{0pt}      \setlength{\parsep}{3pt}
      \setlength{\topsep}{3pt}       \setlength{\partopsep}{0pt}
      \setlength{\leftmargin}{1.5em} \setlength{\labelwidth}{1em}
      \setlength{\labelsep}{0.5em} } }

\newcommand{\squishlisttwo}{
   \begin{list}{$\bullet$}
    { \setlength{\itemsep}{0pt}    \setlength{\parsep}{0pt}
      \setlength{\topsep}{0pt}     \setlength{\partopsep}{0pt}
      \setlength{\leftmargin}{2em} \setlength{\labelwidth}{1.5em}
      \setlength{\labelsep}{0.5em} } }

\newcommand{\squishend}{
    \end{list}  }

\newcommand{\two}{(2)}

\DeclareMathAlphabet{\mathpzc}{OT1}{pzc}{m}{n}

%% file: introduction.tex
\section{Introduction}

Human behaviour is complex and nuanced. Consider how an act as simple as purchasing a cup of coffee involves an intricate spatio-temporal sequence of actions and perception: instructions, clarifications, and feedback weave across language, touch, and visual communicative cues, with the precise timing of each providing yet more information to our interactive partners. If we ever hope to create artificial agents that can participate in similar interactions, we must develop effective ways to evaluate their behaviour in naturalistic settings with humans. 

One obvious approach to evaluating interactive agent behaviour is to leverage a human's judgement during the course of their interaction with an agent. However, this requires a high human cost, both in number of human participants required and in total number of human hours spent, and has no straightforward mechanism to control for human behavioural diversity. The latter problem in particular can result in highly variable metrics if human behaviour is too noisy, or imprecise metrics if human behaviour is not diverse enough. Human behavior is also non-stationary over time, as it can be subtly impacted by agent performance, causing drift. Thus, despite being a "gold standard", the opacity of the online human-agent evaluation setting makes any generated metrics difficult to interpret and communicate, and hence, difficult to optimize for. 

Researchers therefore typically rely on other methods of evaluation, such as validation performance of the agent's optimized objective (e.g. log probabilities for behavioural cloning), or performance during scripted interactions that do not involve humans~\citep{hermann2017grounded}. While such benchmarks are expedient, they may not actually provide insight into how an agent would behave in real-world interactions with humans; scripted interactions in simulated worlds often fail to capture the richness of online human-agent interactions, and, in general, automatic metrics may only loosely correlate with human judgement during online interactive evaluation.

In this work, we explore various ways to evaluate multi-modal interactive agents---including log probabilities measured on behavioural cloning objectives, scripted interactions with automated detection of success, and human judgement during online interactions with agents---and discuss the strengths and weaknesses of each. We then present a novel method called the \name (\acronym), which aligns well with human judgements during online interactions, but is faster to measure, controls for human behavioural diversity, and provides interpretable insights that can guide subsequent research. 

The \name works as follows: first, we identify a set of ``scenarios'' that typify the behaviour we want to evaluate.
In our case the scenarios come from human-human interactions that were recorded within a 3D simulated environment called the Playhouse~\citep{iateam2021creating}.
An example scenario might involve one player asking the other to ``take the ball from the shelf''.
Next, we condition an AI agent by replaying a period of context, forcing its actions to be the same as in the original trajectory.
Upon reaching a ``continuation point'' (which in this example might correspond to the moment that the instruction was given, or the moment when the player approached the shelf), the remainder of the trajectory is generated by sampling actions from the agent's (conditioned) policy.
This agent continuation is then evaluated offline by human annotators as either a success or a failure.
Agents are ranked by the proportion of successes they achieve across a standardised set of many such scenarios.

The \name is a fast and precise evaluation methodology that can be used in any domain to catalyze research in developing agents that interact naturally with humans. Here, we use the \name for the Playhouse domain and present results in the context of MIA~\citep{iateam2021creating}.

%% file: background.tex
\section{Background}\label{background}

While the \name is a domain-agnostic evaluation methodology, we chose to exemplify it in a 3D simulated game environment called the Playhouse~\citep{iateam2021creating}. The Playhouse is a multi-room setting containing furniture and manipulable objects of various shapes, sizes and colors. Interactions are structured around simple language games that unfold between two participants: one plays the role of a ``setter'', who issues instructions and questions to the ``solver''; the solver's job is to follow the instructions and answer questions. 

\citet{iateam2021creating} collected two types of language game data. In the {\em free-form} case, the setter was asked to choose any instruction or question they wanted. In the {\em prompted} case, the setter was asked to improvise an instruction or question based on a given prompt; e.g. if the prompt was ``Ask the other player to bring you something'', the setter might improvise ``Hand me the red object from the shelf.'' \citet{iateam2021creating} used nearly 3 years worth of such data to create a Multimodal Interactive Agent (MIA) using the basic recipe of behavioural cloning and self-supervised learning.\footnote{Throughout the paper we will make reference to a multitude of agents based on MIA. These agents will vary in the subset of data used to train them, their parameter count, hyperparameters, architecture, and so on. We refer to these agents in order to demonstrate how various evaluation methodologies are sensitive to differently constructed agents, rather than to make value judgements regarding specific architectures or algorithms. We direct the reader to Appendix \ref{sec:agent_details} for further details on agent development.}

Creative human behaviour drove both task solving \emph{and} setting in this data, making it impossible to script or even characterise interactions in the Playhouse. Much like how programmatic reward functions were not available for training in \citet{iateam2021creating}, similar functions are not available to \emph{evaluate} agent behaviour in any robust or general sense. \citet{iateam2021creating} instead used two proxy measures (log probabilities, and performance on a small set of scripted probe tasks), and one ``direct'' measure (online human-agent interaction) for evaluation. While these metrics are useful, each is also flawed, as summarized below. 

\paragraph{Interactive Evaluation.} Interactive evaluation comprises judgements by humans when interacting with an agent online and in real-time. For example, in the Playhouse a human might perform the role of a setter, while the agent performs the role of a solver. Interactive evaluation most closely measures the capability of agents to interact cooperatively with humans, which is our objective.

Interactive evaluation, however, has salient drawbacks. In particular, interactive evaluation is expensive and slow since it depends on a scarce resource pool (human participants), and these interactions with humans necessarily occur in real-time. The interactions occur in 5-minute episodes with an average of 3.5 instructions per episode. Each agent is evaluated over $\approx2200$ episodes on average. Interactive evaluation is also difficult to control and interpret since we cannot anticipate which interactions human participants will choose to partake in. This leads to variability of the scores provided by different human participants and by the same human participant across different sessions. Further, humans tend to subtly shift the distribution of instructions that they give according to the competency of the agent in the interaction, giving somewhat easier instructions to agents that are less competent. This behaviour confounds the interpretation of the scores.

\paragraph{Losses and Log Probabilities.} Training losses and validation data log probabilities precisely measure an agent's performance relative to its learning objective (e.g., behavioural cloning of human demonstrations). This information is crucial for understanding the effectiveness of our algorithms and establishing stopping criteria for agent training. Additionally, these metrics are computed online and are automatically available throughout agent training. 

Despite the benefits of measuring losses and log probabilities, there are some weaknesses. In particular, losses and log probabilities often correlate poorly with interactive evaluation, as we demonstrate in Figure \ref{fig:correlations}. Nor is it simple to diagnose particular, relevant behavioural competencies, such as the ability to identify colours or manipulate objects. Log probabilities, which measure prediction losses for target actions, may fail to discriminate how important each action is, especially if training is off-policy~\citep{kumar2022should}. The indiscriminate nature of log probabilities has a large effect because actions are heavily imbalanced, with a large proportion being ``no-op'' actions or other actions less critical to solving the task. For example, in our agent emitting a ``no-op'' (null action) is weighed as heavily as a "grasp"; the latter may be a more important and relevant action in the context of a specific task, but may only happen once in an interaction between a number of ``no-op'' actions. For another simple example, consider an agent solving a T-maze: the only task-relevant action may occur at the choice point of the maze, but this prediction may constitute only a marginal effect on the average log probability.

\paragraph{Scripted Probe Tasks.} Scripted probe tasks comprise researcher-created instructions and reward functions, which can be used to measure success against the instruction. Similar to losses and log probabilities, scripted probe tasks are useful for gaining insights about agents quickly and automatically during agent training. Unlike log probabilities they can be curated to probe specific competencies (such as answering questions, or manipulating objects), and hence are interpretable. 

However, scripted probe tasks are also problematic. In particular, these probes require significant effort to create. They also need to be continuously refined since programming reward functions is difficult. For example, for a ``lift'' task we must specify exactly how high an object should be lifted to be considered a ``success'', and the extent to which this height changes with respect to the object being lifted. Further, programming reward functions is particularly challenging for tasks that require open-ended language emission, such as ``what are you looking at right now''. Overall performance on these tasks also does not correlate well with interactive evaluation, particularly once we reach a certain threshold of competency as shown in Figure \ref{fig:correlations}. It is also impossible to know which individual tasks will correlate well with human judgements prior to expending the effort to create them. Further, scripted probe tasks have poor coverage over the space of ``natural interactions''. Even if a particular task correlates well with interactive evaluation, it can be ``gamed'', rendering it a bad metric. As an example, consider a motor manipulation task that correlates well with interactive evaluation. One can hand-design an agent that does well at this task, perhaps at the expense of all unnecessary abilities to this task, such as the ability to communicate with language. An agent without an ability to emit language will be severely limited in its ability to interact with a human in most respects, despite its narrowly gauged motor manipulation performance.

%% file: sts.tex
\section{The \name}

\begin{figure}[ht]
    \centering
    \includegraphics[width=.9\textwidth]{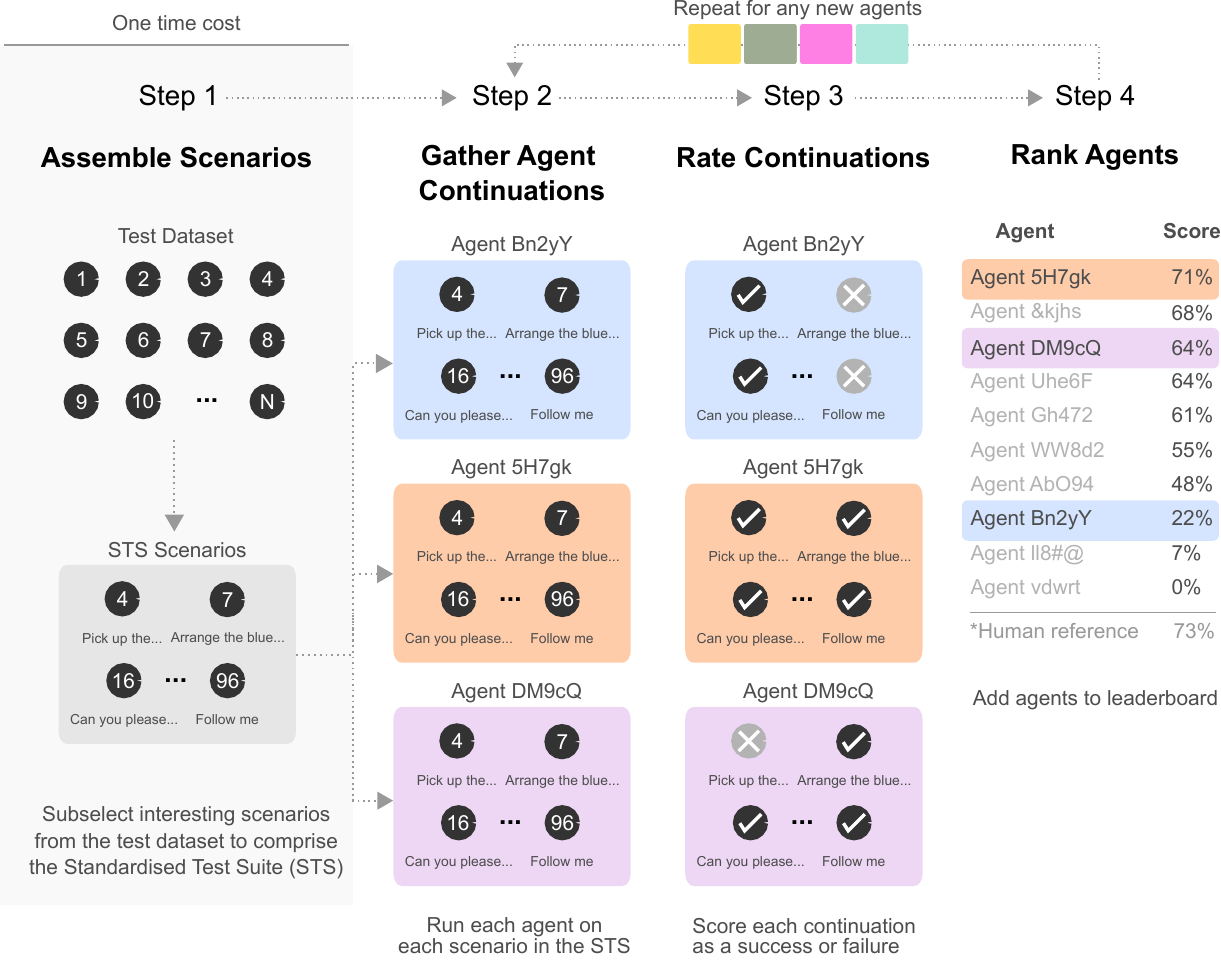}
    \caption{\small{\textbf{The \name workflow.} The first step in the \name is to compile a list of interesting interaction scenarios. In our case we leveraged data that was already gathered and randomly assigned to the test split used for calculating metrics during behavioural cloning experiments, and sampled interactions that had rough coverage over the kinds of interactions that take place in the Playhouse (e.g., lifting objects, or answering questions about object properties). Crucially, though this step requires human input, it is a one-time cost. We then generate agent continuations offline on this chosen subset, send them to humans to rate as success/fail, and then use these ratings to rank agents.}}
    \label{fig:sts_flow}
\end{figure}

We created the \name (\acronym) as a new system to evaluate multimodal interactive agents. The \acronym \space uses a curated set of scenarios to test core competencies of agents. Figure \ref{fig:sts_flow} shows how the \acronym \space works at a high-level: we first curate a set of human-human interactions that we believe represent interesting examples of interactive behaviour and provide good coverage of desired competencies. We then force agents to act exactly as prescribed by the humans who produced these trajectories. Upon reaching a predefined ``behavioural continuation'' point, agents are made to act on their own for a fixed amount of time. Agent continuations are evaluated offline, by humans, and the resulting evaluations are used to rank agents.

The \acronym \space has several key features that overcome the disadvantages of existing evaluation metrics, making it a suitable complement to many evaluation ecosystems. In particular, the \name is:
\begin{itemize}
    \item \textbf{Direct}: it is similar to interactive evaluation, and, in contrast to log probabilities, the \acronym ~measures interactive behaviour directly.
    \item \textbf{Comprehensive}: its scenarios can easily be gathered to cover a wide range of categories and difficulty, unlike scripted probes which require substantial effort to create.
    \item \textbf{Categorized}: the chosen scenarios can be grouped as the researcher desires, thus increasing the interpretability of agent performance on the metric. 
    \item \textbf{Stationary and low variance}: it is trivial to generate multiple behavioural continuations per scenario to decrease measurement error or, similarly, to gather multiple human evaluations per scenario.
    \item \textbf{Extendable}: the gathering of scenarios is a one-time cost, and users can incrementally gather new scenarios to include in the suite. 
    \item \textbf{Aligned with human judgements}: \acronym \space rankings correlate well with rankings from interactive evaluation.
    \item \textbf{Low cost}: evaluating an agent on the \acronym \space requires a fraction of the time used for online interaction since ``scrubbing'' through prerecorded videos to evaluate them for success or failure is faster than an online interaction in which one needs to conceive of an instruction and engage with the agent.
\end{itemize}

The following sections will describe each step of the \acronym \space in more detail, outlining some of the design decisions and possible points of departure for those who wish to iterate on this basic design.

\subsection{Constructing the \acronym \space}\label{sec:constructing_sts}

The first key stage of the \acronym \space is creating a set of \textbf{scenarios} that is comprehensive and represents naturalistic interactions. This step can be performed in a few ways depending on the research agenda.

One could first determine interesting or useful categories of behaviour (such as lifting objects, or answering questions about their characteristics), and then search for scenarios that cover a range of difficulty and types of interaction within each category. Finding scenarios for each category can be accomplished through  data-mining, or by recording new data designed to probe the identified categories. However, there may not be an easy, automatic way to sift through datasets to find scenarios, so this approach may also require more time and effort. Moreover, this approach of identifying scenarios towards specific categories may cause the metric to stray from interactive evaluation if the selected categories are underrepresented in interactive evaluation data, even if the resulting metric better captures certain research goals. 

Another approach consists of first identifying a source of data and then creating categories based on inherent characteristics of this data. For example, one can take a dataset of human-agent interactions (gathered through interactive evaluation, for example), and then bucket each sample based on whether the interaction was rated as successful or not by the human participant. The \acronym \space scenarios could then comprise samples from the ``unsuccessful'' bucket, for example, or some weighted mix of the successful and unsuccessful buckets. One advantage to this approach is that scenarios can be quickly and automatically gathered and grouped. However, categorizing the scenarios will ultimately be limited by the exposed characteristics of the data. The resulting set of scenarios also more directly represents the underlying distribution of training data (though the episodes comprising the scenarios are not used during training themselves) and the data collected during interactive evaluation, leading to a higher likelihood of correlating strongly with interactive evaluation.

For our initial set of scenarios we used a combination of the two approaches described above. We chose to closely align our categories with the prompt categories we use in interactive evaluation, which we believe comprises a broad spectrum of competencies that we hope all agents would possess. Thus, some of the 16 categories were congruent with the prompts given to setters (allowing us to quickly filter the data to choose scenarios) while some were not congruent (forcing us to more intensively mine our data). We sifted through a month's worth of human-human prompted data to compile ten scenarios for each category. Importantly, the data from which we selected the scenarios was not used for training. Finally, we refined our scenario choices by ensuring there was a range of difficulty within each prompt, as judged by human intuition. For example, challenges could come in the form of the instruction given, the objects mentioned, or the layout of the room. We also made sure to include easy scenarios for each category, in order to differentiate agents across the spectrum of capabilities. 

\begin{figure}[ht]
    \centering
    \includegraphics[width=.85\textwidth]{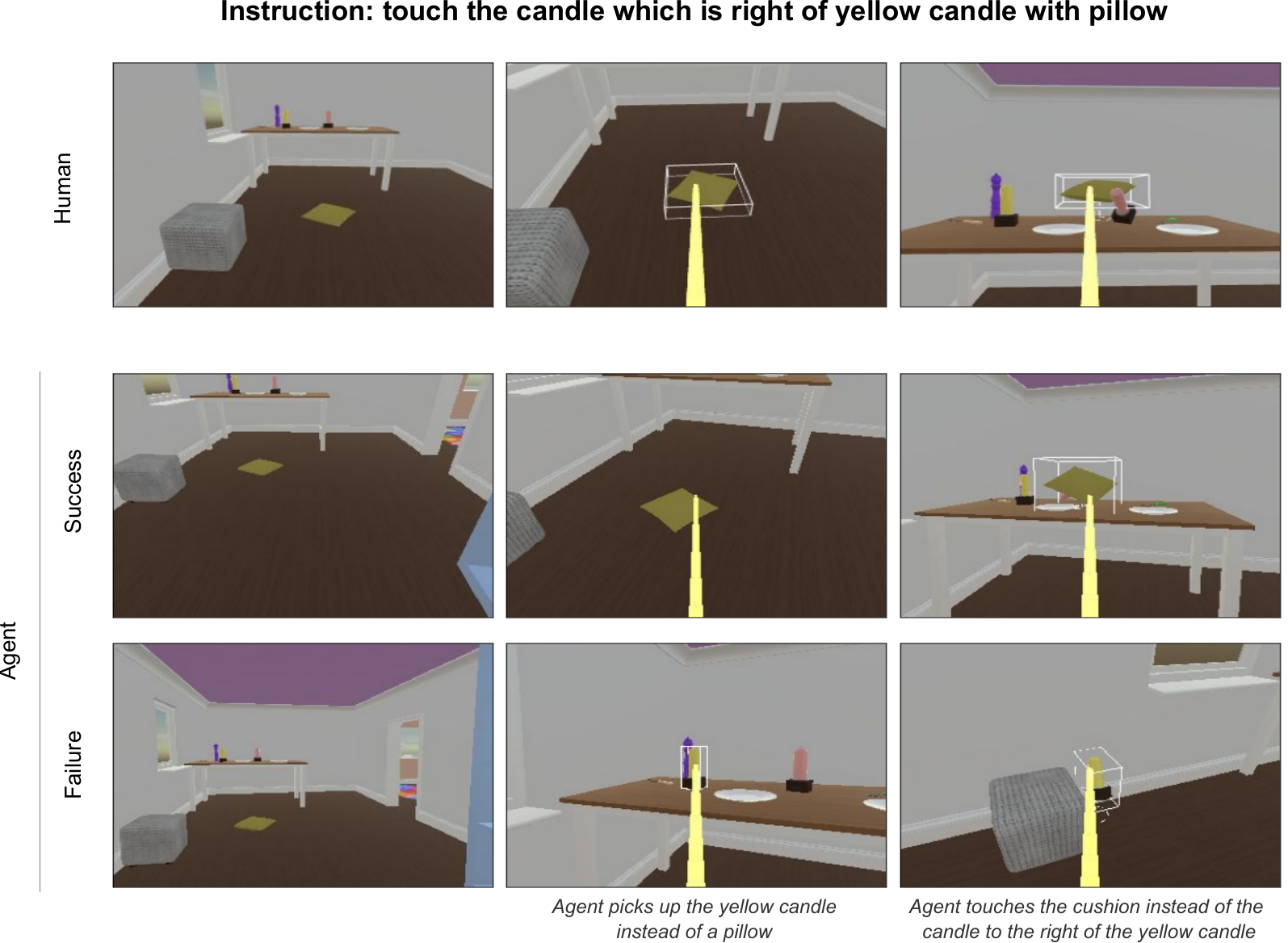}
    \caption{\small{\textbf{Examples of instructions and key frames from the \acronym.} The top row depicts three extracted frames from successful human behaviour when instructed to touch a particular candle with a pillow (human judges, on the other hand, have access to a full video depiction of behaviour which they can skip through indiscriminately). The bottom two rows depict an agent success and failure. Notably, despite having failed, the agent nevertheless exhibits purposeful and intricate motor behaviours: it picks up an object (albeit an incorrect one) and navigates through its environment to touch this object with another.}
    }
    \label{fig:continuations}
\end{figure}

\subsection{Generating Behavioural Continuations}

Once a fixed set of scenarios has been established, we can generate \textbf{behavioural continuations} offline. Each scenario has an associated context period, takeover point, and continuation length. During the context period, the agent acts as an observer; observations and actions are taken from the trajectory in the dataset -- not from how the agent would have acted in those scenarios. At the takeover point, the agent is left to continue taking actions in the environment for the entirety of the continuation length. This sequence of actions and observations, from the start of the context to the end of the continuation, is recorded as a new episode. For example, in the scenario presented in Figure \ref{fig:continuations}, the agent is left to act after receiving the instruction ``touch the candle which is right of yellow candle with pillow''. The successful agent continuation looks very similar to the human trajectory, with the agent picking up the pillow in front of the table and touching the correct candle. An unsuccessful continuation also yields useful insight -- here, the agent understands the ``touch'' and ``yellow candle'' portions of the instruction, but does not understand the instruction as a whole -- instead picking up the yellow candle (wrong!) and touching it to a nearby cushion (wrong!). This process of generating continuations can be repeated as many times as desired in order to average out trial-by-trial noise.

\subsection{Rating the Behavioural Continuations}\label{Rating}

The recorded continuations are then sent to human annotators, who watch videos of the continuations and place a single marker at the moment of success or failure, called an annotation. Annotators are instructed to place the marker on the first instance of success (if there are multiple) or the first instance of failure (if there are multiple, or if the agent fails before it succeeds). This process of sketching a binary success/fail is significantly faster than playing a whole episode with an agent in interactive evaluation, making the \acronym \space more time-efficient. Overall, the number of human hours required to evaluate an agent using the \acronym \space is, empirically, over $6x$ faster on average than interactive evaluation. However, humans are not perfect, and sometimes place markers incorrectly (e.g. place a success instead of a failure marker, or place the marker at the wrong time), as detailed in Appendix \ref{sec:human_errors}. Despite these errors, humans still achieve close to $90\%$ balanced accuracy on rating continuations as measured on a set of reference episodes, so any variance caused by errors in rating is small.

\subsection{Ranking Agents}\label{sec:ranking_agents}

\begin{figure}[ht!]
    \centering
    \includegraphics[width=.65\textwidth]{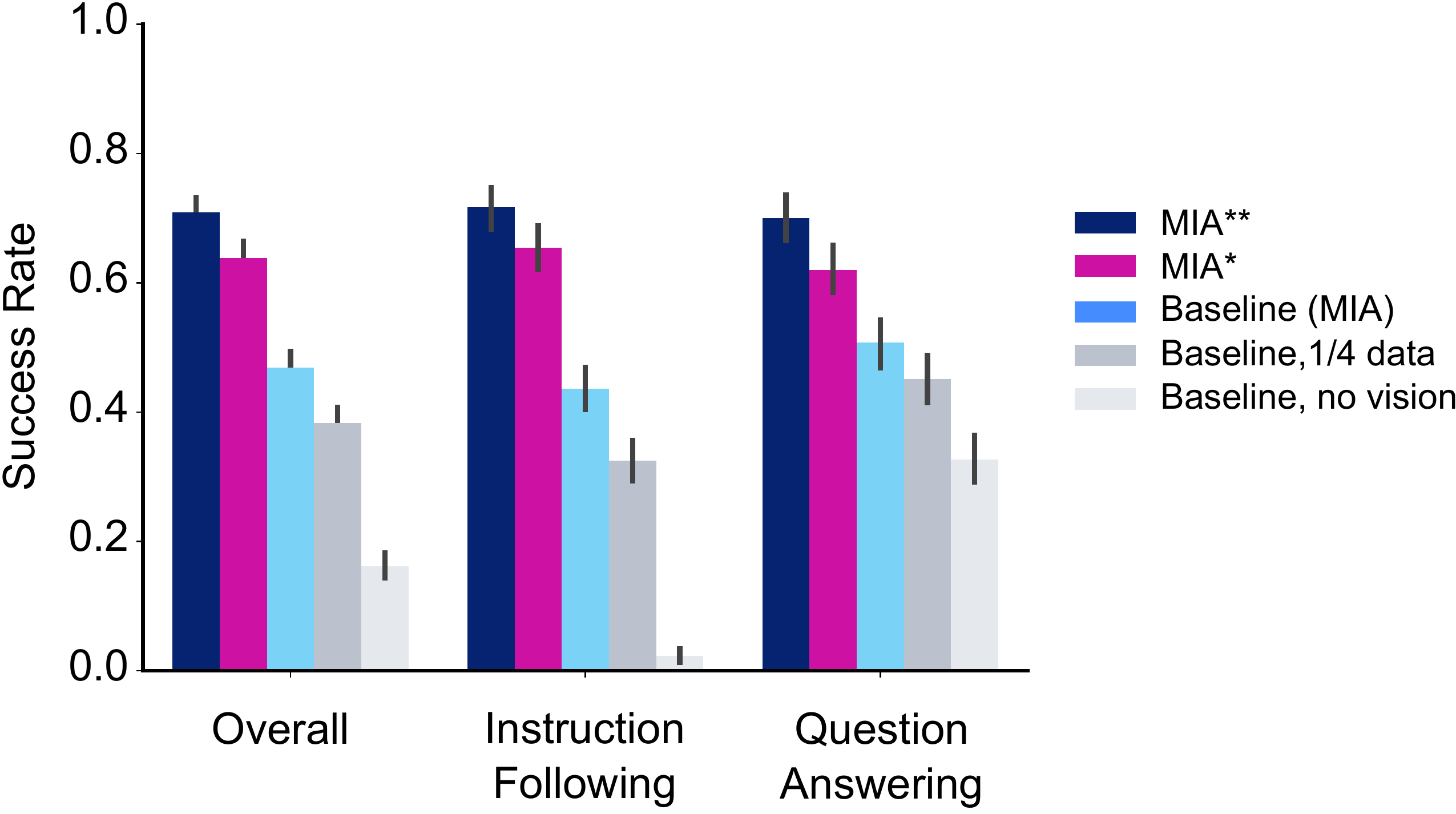}
    \caption{\small{\textbf{Example agent ranking}. The \acronym \space enables ranking by both overall score and by any desired sub-category. In our Playhouse setting, we found it useful to categorize scenarios into those that mainly involved motor action (``Instruction Following''), such as lifting objects, and those that mainly involved language emission (``Question Answering''), such as responding to whether an object exists in the room. Depicted are a subset of agents based on the MIA baseline from \cite{iateam2021creating}. Error bars represent standard error across all continuations in each category, with ten continuations per agent per scenario yielding 1,600 continuations per agent overall. For more details on the agents ranked here, see Appendix \ref{sec:agent_details}.
    }}
    \label{fig:agent_performance_comparison}
\end{figure}

Once all behavioural continuations for an agent are annotated, we can synthesize the results into a single metric -- the percentage of continuations in which the agent succeeded. This number can be easily used to rank agents and measure progress. Figure \ref{fig:agent_performance_comparison} shows the overall results of several key agents, including MIA from~\cite{iateam2021creating}. The function of a truly robust evaluation system for multimodal interactive agents, however, is not only to provide a good measure of performance but also to facilitate a deeper analysis of agent capabilities. Thus, in comparison to just looking at a single metric, we can better interpret the underlying results by watching videos of individual continuations for specific scenarios and viewing per-category performance, which is also demonstrated in Figure \ref{fig:agent_performance_comparison} by presenting a breakdown of the overall score into a score on instruction-following scenarios and a score on question-answering scenarios. By breaking down the overall score into categories, some interesting observations emerge; for example, the ``no vision'' baseline performs reasonably well at question answering tasks, presumably because it leverages knowledge of the base level statistics of answers in the data and because some questions have a limited number of potential responses (e.g. yes/no questions) making it easy for the agent to guess correctly. On the other hand, it fails nearly completely at instruction following, where a similar strategy cannot be exploited. Calculating the overall pass rate is the simplest possible statistic, and adopting an improved statistic that penalises agents that guess is left for future work. For a more detailed breakdown of performance by category, see Appendix \ref{sec:categorization}.

%% file: results.tex
\section{Results}

In this section, we provide direct comparisons of all the evaluation metrics in our ecosystem, we show that the \acronym~ provides a useful signal for comparing agents and correlates best with interactive evaluation, and lastly, we demonstrate the progress made as a result of the \acronym.

\subsection{Assessing Evaluation Metrics}

Though previous evaluation metrics that we explored, as defined in Section \ref{background}, are each still useful, they have problematic qualities that made our evaluation practices suboptimal.

\begin{figure}[ht!]
    \centering
    \includegraphics[width=0.9\textwidth]{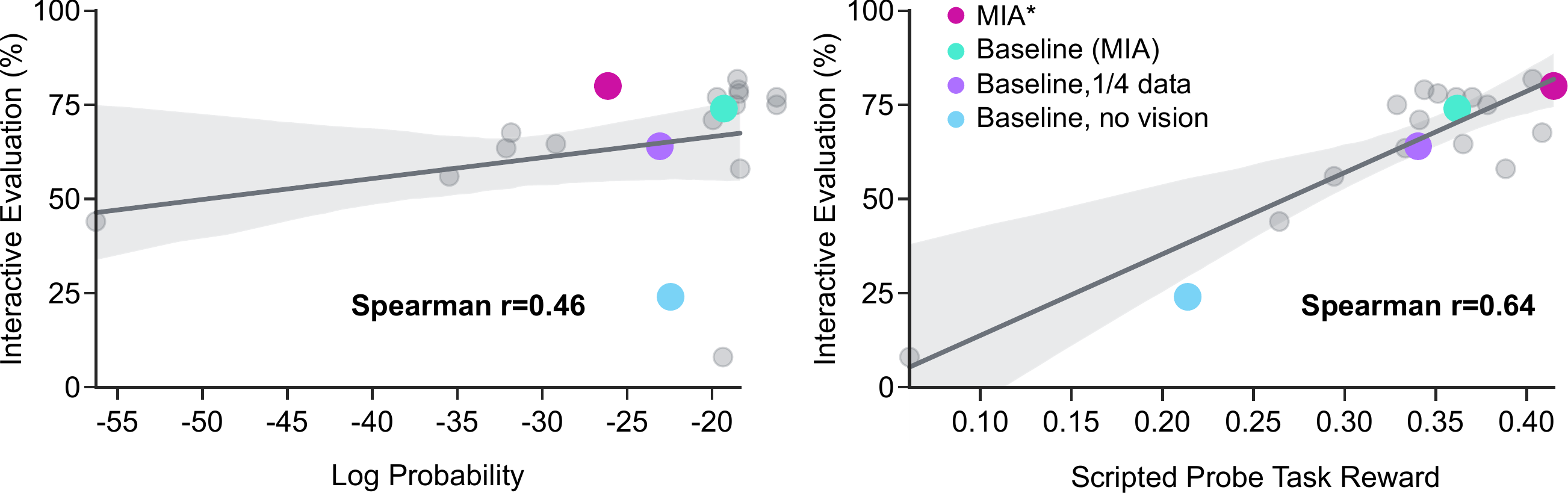}
    \caption{\small{\textbf{Correlations between evaluation methods.}
    Interactive evaluation between humans and agents is in many ways a gold-standard, but is also a slow and expensive metric. Here, each point represents an individual agent, and we highlight several important baselines. Other metrics are fast to compute, such as log probabilities, but might not correlate well with interactive evaluation (Spearman r=0.46, \textit{p}=.045 for our agents). Scripted probe tasks are more interpretable than log probabilities, as they assess particular behavioural competencies directly, but are expensive to construct and also don't necessarily correlate strongly with interactive evaluation (Spearman r=0.64, \textit{p}=.003 for our agents). Moreover, each metric is differently sensitive across its dynamic range; for example, many high-performing agents cluster towards the limit of achievable log probabilities, making log probabilities a sensitive and informative metric mainly for poorly performing agents.
    }}
    \label{fig:correlations}
\end{figure}

Though interactive evaluation most closely aligns with the ultimate goal, to create agents that can interact cooperatively with humans, using these interactions to evaluate agents is insufficient, as this form of evaluation is expensive, time-consuming, and difficult to interpret (among other disadvantages). Further, Figure~\ref{fig:correlations} shows that automatic metrics such as losses and log probabilities and performance on procedural levels are only roughly correlated with interactive evaluation and are, therefore, poor proxies.

Thus, there are both qualitative and quantitative problems with past evaluation methodologies that created a need for a new evaluation system. We have explained how the \acronym \space provides additional affordances such as speed and interpretability as compared to interactive evaluation, but the \acronym \space must also correlate well with interactive evaluation in order to satisfy the criterion of aligning with human judgements of agents and providing information beyond the automated metrics (log probabilities and scores on scripted probe tasks) already available.

\begin{figure}[ht!]
    \centering
    \includegraphics[width=0.9\textwidth]{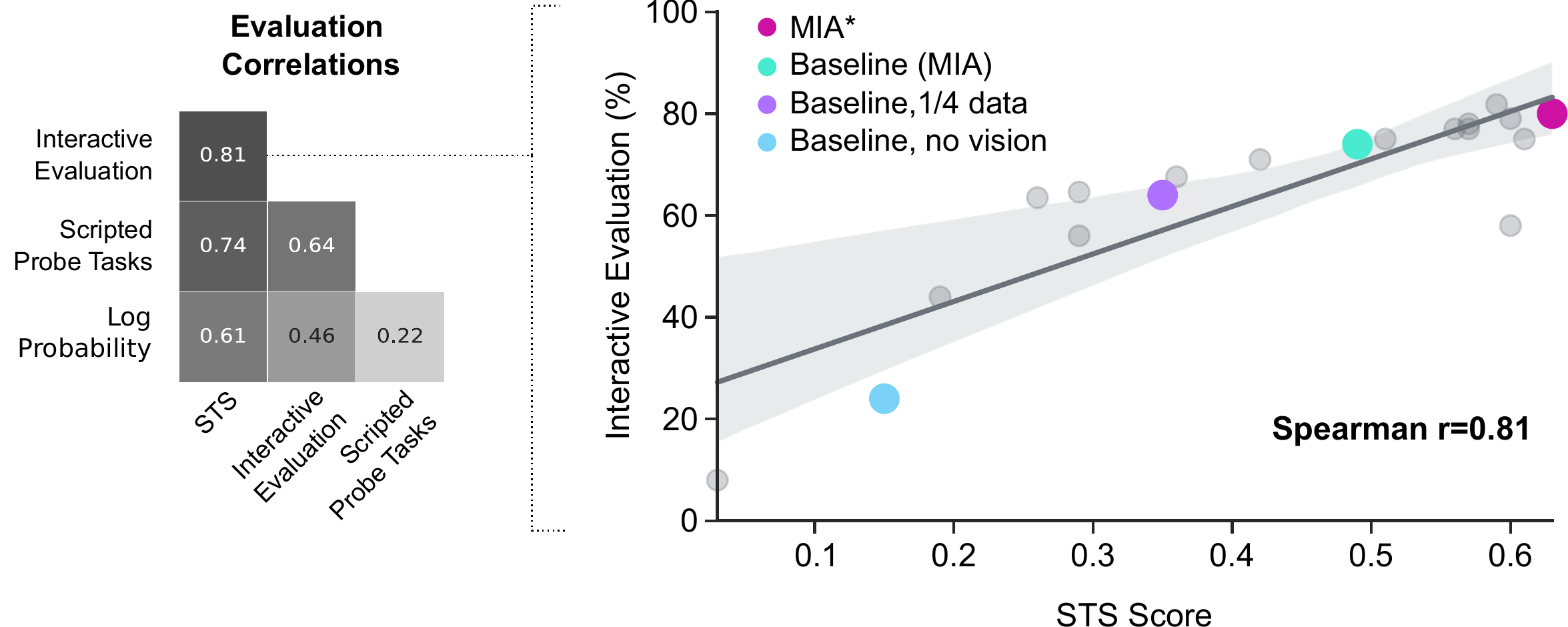}
    \caption{\small{\textbf{The \acronym \space correlates best with interactive evaluation.} Unlike log probabilities and scripted probe tasks (see figure \ref{fig:correlations}), the \acronym \space is both sensitive across its dynamic range and correlates well with interactive evaluation. Moreover, this correlation and sensitivity can be explicitly manipulated during the first stage of constructing the \acronym; scenarios can be chosen or filtered based on their effects on these properties, thus making it possible to construct an \acronym \space that is as precise and accurate as one desires. \textit{P} values are significant at <.05 except for the correlation between log probability and scripted probe tasks where \textit{p}=.35.
    }}
    \label{fig:sts_vs_ha}
\end{figure}

Figure \ref{fig:sts_vs_ha} shows that \acronym \space scores correlate strongly with interactive evaluation, with a Spearman rank correlation value of $.81$, \textit{p}<.001. Interestingly, the \acronym \space score also more strongly correlates with procedural scores and log probabilities than interactive evaluation does. This suggests it captures both important aspects of natural interaction, like interactive evaluation, and specific quantitative metrics that are optimized during training, like the probability of matching human behaviour.

Each of the metrics above plays an important role in the ecosystem of evaluating interactive agents: log probabilities and scores on scripted probe tasks are automatic and online, providing a useful immediate signal for whether an agent will be particularly bad. However, once we reach a certain threshold, this signal becomes too noisy to trust. Interactive evaluation, though most representative of the ultimate goal scenario, is too costly and high variance to fuel research progress. In contrast, the \acronym \space provides a compromise that is faster than interactive evaluation and more representative than automated metrics -- a method for simulating naturalistic interactions to create a faster and more controlled evaluation metric.
 
\subsection{Accelerating Research Progress Using the \acronym}\label{sec:progress}

\begin{figure}[ht!]
    \centering
    \includegraphics[width=.6\textwidth]{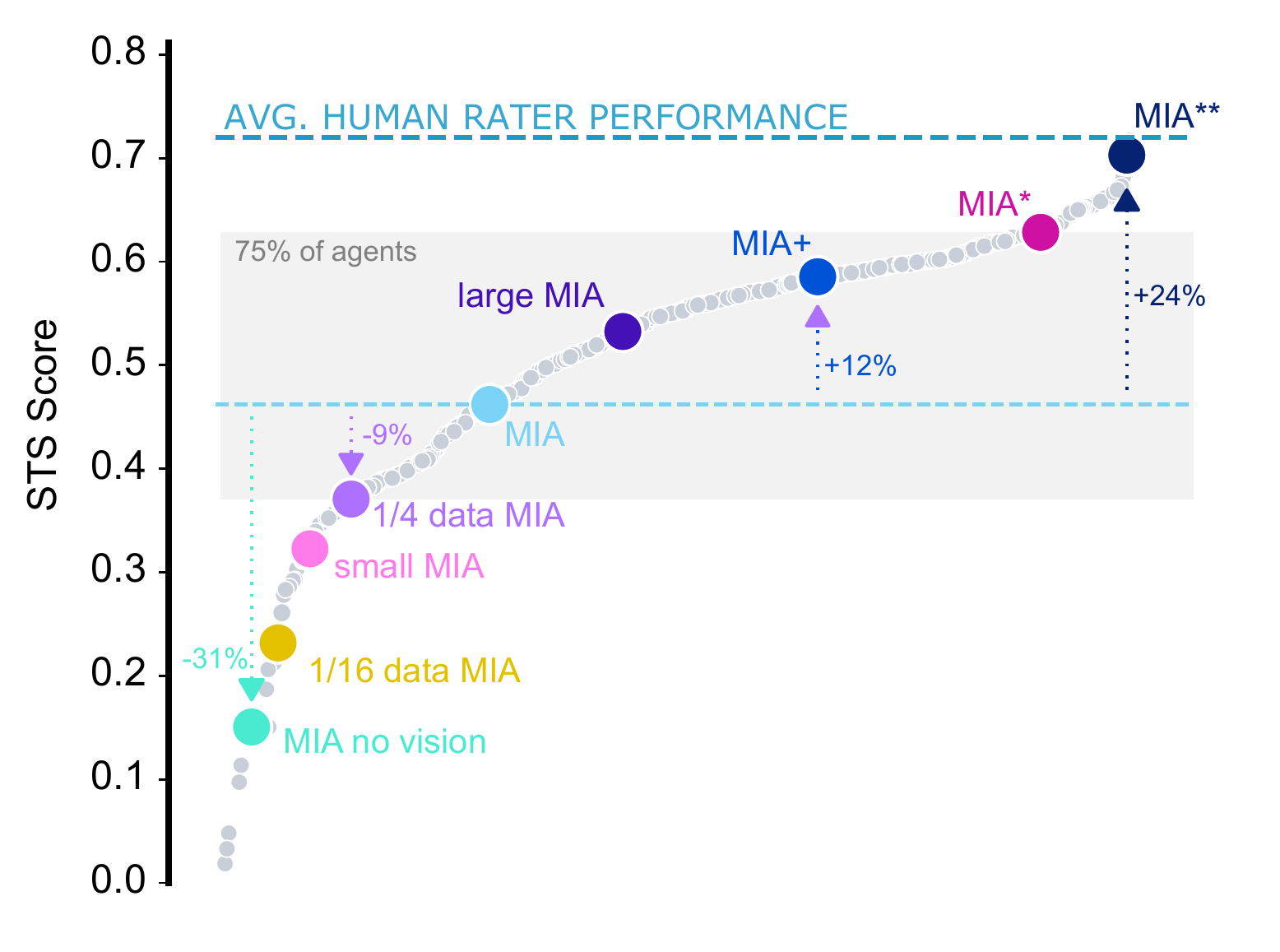}
    \caption{\small{\textbf{Agent progress on the \acronym, ranked by performance.} Since using the \acronym, we've developed a number of agents with variable performance, exhibiting a dynamic range of approximately 70\% in score for this particular scenario set. Each point represents an agent that we trained since launching the \acronym. The \acronym \space reveals a number of expected effects, such as the impact of scaling both data and model size (both of which improve performance), as well as the effect of ablations (such as removing vision from the agent). While the average human annotator is roughly 73\%, expert humans can achieve close to 100\% on this suite (data not shown). While agent performance has not yet plateaued, breaking down performance results by category (see Figure \ref{fig:agent_performance_comparison}) could reveal whether remaining potential performance gains will be due to, for example, improved performance in a small set of scenarios, or marginal improvements across the suite, each of which could uniquely motivate agent development. The superscript symbols (+, *, **) refer to unique variations of the baseline MIA agent (by, e.g., changing parameter count, architecture, or data; see Appendix \ref{sec:agent_details}).}
    }
    \label{fig:agent_progress}
\end{figure}

The ability to efficiently and reliably compare changes to an agent's data, size, architecture, or algorithm has enabled us to make swift progress as demonstrated in Figure \ref{fig:agent_progress}. Because the majority of the \acronym~ workflow is automated and the \acronym~ is more time efficient as described in Section \ref{Rating}, we can use $25\%$ fewer human annotators and evaluate $5.5x$ as many agents per month as compared to interactive evaluation. Lastly, the increased interpretability of evaluation as discussed in Section \ref{sec:ranking_agents} has greatly enhanced our understanding of our agents and their capabilities. We have evaluated over 500 agents and achieved a $24\%$ increase in \acronym~ score as a result of improvements along many dimensions. 

Though we appear to have reached average human annotator performance, we believe that theoretically both humans and agents have the ability to reach $100\%$ accuracy on this set of scenarios. There are two sources of human error resulting in a lower average performance for humans than expected: incorrect completion of the task and incorrect rating of the behavioural continuation. Some humans are less adept at the controls and thus fail to complete the task correctly, and at the same time, some humans are particularly strict when judging success, marking other human's behavioural continuations as failures for minor offenses. Further, humans can make mistakes as well and misread instructions or mislabel a continuation. However, practice can enable humans to get better at these tasks, and there are some human subjects who show much higher than average performance. For more details on the breakdown of human errors, see Appendix \ref{sec:human_errors}. Thus, we believe that there is potential for agents to continue to demonstrate improvements on this set of scenarios.

%% file: additional_features.tex
\section{Additional Features}

\subsection{Expanding the \acronym}

One of the most powerful features of the \acronym \space is its ability to grow over time while maintaining its core benefits as research goals and agent performance evolve. As we continue to make progress at a steady rate, we are already reaching approximately human performance on the initial suite of scenarios; this achievement is not a limitation of the test suite, though, but rather a testament to its ability to unlock significant progress. 

There are several key features incorporated in the design of the \acronym \space that enable easy expansion. First, we can expand the \acronym \space in batches to create versioned benchmarks, which are essential to maintaining a consistent standard. Rather than continuously updating the suite of scenarios, we can expand the \acronym \space incrementally and continue to generate scores on previous versions in order to be able to fairly compare different agents over time. Second, we utilise categorisation and tagging to aid in interpretability as the suite grows. Each scenario is tagged with a category so we can easily generate breakdowns of results. Tagging functionality goes beyond just categorisation, however -- we can tag scenarios with their specific version in order to maintain a reference to each version as the suite expands, and individual researchers can also submit their own tags in order to view personalised results breakdowns.

\subsection{Measuring Additional Characteristics of Agents}

In addition to enabling us to iterate quickly and reliably on agents, the \acronym \space also allows us to analyse properties of agents beyond just success or failure and to evaluate agents that scale beyond what is feasible in real-time. For example, we can measure characteristics such as time to completion and consistency, as shown in Appendix \ref{sec:ttc} and \ref{sec:consistency}. For time to completion, it is generally accepted that agents that succeed faster are better-- perhaps they are more goal-directed, less clumsy, or better at navigating their environment. In fact, we see this hold true for many of our agents, as demonstrated in Appendix \ref{sec:ttc}. Thus, being able to compare agents' speed at succeeding at tasks is useful for gaining further insights into an agent's capabilities and predicting agents with whom it would be qualitatively better to interact. 

Another characteristic that we can measure is consistency, which comes in two forms: consistency within a continuation and consistency across continuations (e.g. when faced with the exact same scenario, how often does the agent succeed?). We can measure the first type of consistency, consistency within a continuation, by creating scenarios that test specific attributes that we hope agents to have, such as logical consistency. The latter is a measure of average success on a given scenario, as in our current procedure, we generate ten behavioural continuations per scenario for each agent. Although the physics simulator underlying the Playhouse environment is not entirely deterministic, the stochasticity across continuations comes primarily from the agent's policy. Thus, for agents without deterministic action sampling, tracking the percentage of continuations that were successful per scenario is an interesting metric. For particularly poor agents, this insight gives us a sense of how good an agent is at randomly guessing. Similarly, we expect (and, indeed, observe) better agents to succeed more consistently, as demonstrated in Appendix \ref{sec:consistency}. Note that this form of consistency is also useful to measure the difficulty of a scenario-- harder scenarios are completed less frequently even by good agents, and easier scenarios are more consistently succeeded on by a wider range of agents. We can then use the distribution of agent consistency to ensure that we have a range of difficulty across scenarios and categories.

Lastly, because the \acronym \space is offline, this evaluation is not constrained by real-time interactions. Though we ultimately want to produce agents that are indeed fast enough to run in real-time (30Hz), it is easier test the limits of scaling to create better agents first and then later research ways to make agents faster rather than limit ourselves to real-time interactions under existing infrastructure. In fact, while pushing the boundaries of scale, we have demonstrated that agent performance directly correlates with scale (see Appendix \ref{sec:fps}), and our best agent, MIA**, is actually too large to run interactive evaluation on at all. Follow up work will aim to recover real-time agent inference whilst maintaining performance.

%% file: limitations.tex
\section{Limitations}

Though the \acronym \space has significant advantages over other evaluation metrics, and provides a good compromise between full interactive evaluation and automatic metrics, there are some limitations to this system worth noting. 

First, curating scenarios requires time and effort. Though some of the process can be largely automated by selecting scenarios from agent failures in human-agent interactions, there are some cases where finding particular scenarios that test a specific skill aligns better with research goals. However, curating scenarios is a fixed cost performed only when you want to create a new version and is thus justifiable. 

Because the scenarios selected are just a subset of all possible scenarios, there is also the issue that sensitivity is not necessarily linear across its range as the scenarios comprising the suite vary in difficulty. Consider a situation where progress from $40\%$ to $60\%$ is driven by maximising performance on an easy/medium set of episodes, but then performance stalls above $60\%$ as the remaining scenarios are a significant jump in difficulty for which there hasn't been any positive signal throughout agent development. To overcome this issue, we must ensure that there are a sufficient number of "medium" level scenarios, and be vigilant about identifying plateaus in progress that may indicate the need for a new version.

Additionally, the behavioural continuations paradigm requires the ability to reinitialise the environment at the continuation point, such that it is in precisely the same state as in the original trajectory from which the scenario was derived, which is likely to be intractable for non-simulated environments. Implementing and maintaining the serialisation mechanism underlying this capability also required significant engineering investment, especially as the environment became more sophisticated over time.

Behavioural continuations are generated offline, which means there is no human player available to provide actions on behalf of the setter after the continuation point. Although we could continue to replay the setter's actions from the original trajectory, these actions are unlikely to make sense given the new actions produced by the agent. We therefore force the setter to take no-op actions after the continuation point, which precludes behaviours that require active participation from the setter after the continuation point (such as two-way dialogue, or the solver handing over an object to the setter). We can approximate multi-turn interactions by creating multiple scenarios that each consist of a single turn in the original human interaction. For example in the case of dialogue we could have a series of scenarios that each test whether the agent's next utterance is appropriate given the preceding dialogue from the original trajectory. However this is probably not a very realistic test of how well an agent would perform in a real multi-turn interaction, since agents may be less likely to produce appropriate actions if allowed to deviate from the original human trajectory. Another possible solution would be to use pretrained agents to provide actions on behalf of the setter in a closed-loop fashion, allowing the setter to respond to the agent's behaviour within the continuation.

Furthermore, as discussed in Section~\ref{sec:progress}, humans are not perfect and can make mistakes when judging the success of a continuation, leading to variation and errors in results. To track annotator accuracy, we maintain a set of reference episodes with true labels that are randomly served to human annotators. We can then use this accuracy to help measure variance in our results and ensure that our quality is not degrading.

Lastly, the \acronym \space  is difficult to ship to the external research community-- the environments, data, and annotation tools are not open-sourced, and the system still requires human annotators who operate on platforms coupled with internal infrastructure. We hope, however, that the design of this system can inspire similar evaluation systems for other researchers in the community.

%% file: related_work.tex
\section{Related Work}

The field of intelligence testing has a long and socially controversial history, beginning its modernity with Spearman's studies \citep{spearman1904general}, which prominently demonstrated that multiple independent measures of performance tend to be correlated in individuals. As a corollary, Spearman showed that sufficiently many measures of a person's performance predict a second, unrelated group of distinct measures. 

Surprisingly, the exact content of these measures was not itself important: the measures could include simple sensory discrimination tasks (for example distinguishing two tones), as well as more complicated tasks that require education, such as mental arithmetic \citep{duncan2010intelligence}. After Spearman's result, more homogeneous, and \emph{standardised} tests (such as Raven's Progressive Matrices \citep{raven1938raven}) were developed that could be administered more easily and quickly. In our work, interactive evaluation is analogous to Spearman's early measure of multiple independent tests: a set of independent interactions with multiple people is predictive of performance in further interactions (that is, it predicts agent performance in free-form interactions with the broad human population). Continuing this analogy, the \name serves the same role as homogeneous, standardised tests like Raven's Progressive Matrices, which can be administered efficiently.

In Artificial Intelligence, Turing himself proposed that the ultimate test of machine intelligence is its ability to interact with human evaluators \citep{turing1950computing}. This suggestion spawned a long-standing tradition of interactive evaluation, which, perhaps because of the non-stationarity of human interactive evaluation and the historical primitiveness of AI behaviour, has primarily stood to the side of mainstream machine learning research \citep{loebner2009hold}. However, in subfields where AI performance is closest to human capability, human interactive evaluation is the definitive standard to prove an AI's capability \citep{silver2016mastering}.

The subfield of natural language processing is notable for its sophisticated battery of evaluation benchmarks. Along with the preponderance of high-quality text data and the simplicity of scaling language models, these benchmarks have helped steer the field toward rapid progress \citep{brown2020language, rae2021scaling}. Recently, with the arrival of highly capable language models, human evaluation has become a crucial tool, allowing the dynamic evaluation of models as they improve \citep{kiela2021dynabench, thoppilan2022lamda}. These methods are complementary to more static benchmarks like SuperGLUE \citep{wang2019superglue}. Similar to developments in language modeling, we find that validation data log probabilities are important diagnostics but often do not always correlate well with human evaluations, necessitating other standards \citep{papineni2002bleu, banerjee2005meteor} to judge the quality of AI decisions when human evaluations are expensive. We likewise use scripted probe levels as an important adjunct during model training to assess under- and over-fitting.

The development of embodied, grounded language agents \citep{hermann2017grounded} that can engage in open-ended interactions is still relatively immature. Some work has been done on embodied agents but with no or limited language capabilities \citep{DBLP:journals/corr/abs-1910-03655, DBLP:journals/corr/abs-1809-00786}. Other work has been done on grounded language agents but their capabilities are restricted to a limited task space \citep{bara-etal-2021-mindcraft, thomason:corl19, TEACh22}. However, recent work in simulation and robotics using human demonstration data has pushed the capabilities of grounded language agents toward open-ended, language-conditioned interaction \citep{lynch2020learning, abramson2020imitating, iateam2021creating,padmakumar2021teach}. To our knowledge, this work presents the first careful study of a metric that evaluates open-ended interaction in a standardised way. We hope that the design principles underlying the \acronym \space will allow researchers to design similar evaluation benchmarks that can accelerate progress in this rich area.

%% file: conclusion_eval.tex
\section{Conclusion}

Fast and effective evaluation that measures progress on human judgements is at the heart of recent advances in machine learning. Consider image classification, where progress has fundamentally been driven by collecting human annotations of images. In this sense, MNIST, CIFAR-10, and ImageNet underlie progress on developing effective algorithms and architectures for perceptual tasks. In these supervised learning regimes, once the original human judgements are collected, creating evaluation mechanisms that are straightforward and automatic is relatively easy. In other cases, it may not be possible to have a single metric to track progress. For example, in language modelling, image captioning, and image generation, researchers have shown that metrics such as BLEU and FID correlate better with human judgement than the losses used for training. In the case of naturalistic interactions in more complex environments, evaluation is even more challenging, and the lack of effective metrics that correlate well with human judgement is likely to hold back progress in this area of research.

In this work, we assess existing metrics used to evaluate multimodal interactive agents, identifying a need for a new system that is faster and more controlled than open-ended interactive evaluation but that correlates better with interactive evaluation than existing automatic metrics. We then present the design for such a system called the \name that utilises behavioural continuations on a fixed set of representative scenarios to rank agents. We show that when implemented for MIA-style agents in the Playhouse environment, the \acronym \space indeed correlates best with interactive evaluation while reducing human cost and time required for evaluating agents. Lastly, we demonstrate that good evaluation systems in practice greatly accelerate research progress.

There are several areas of future work that remain for the \acronym \space in order for the system to reach its full long-term potential. One challenge is expanding the set of scenarios while maintaining low human cost. We are currently exploring training models to automate the human annotators by predicting the success of a continuation of existing scenarios, which would allow us to create a new version of the \acronym \space without increasing the cost, and this cycle is repeatable ad infinitum. Another challenge is finding ways to further automate the scenario selection process, such as harvesting agent failures from existing data collections. Finally, we must continue to monitor the contents of the \acronym \space to ensure that the metric aligns with our goals and maintains a strong correlation with interactive evaluation.

Though this paper is a step towards creating metrics that the community can use by providing a model for such an evaluation system, we cannot yet provide a portable solution for other researchers to use, and note that this method is only possible in a simulated environment that we can reinitialise to specific states from prerecorded trajectories. This method still requires humans in the loop, though we believe that there is a straightforward path toward automating the entirety of the process. That said, the methodology that the \acronym \space employs is domain-agnostic, and other researchers can recreate such a system in any domain to align with their own goals.

%% file: authors.tex
\newpage
\section{Authors \& Contributions}

\textbf{Josh Abramson} contributed to agent development, data and tasks, engineering infrastructure, and as a technical lead. \\
\textbf{Arun Ahuja} contributed to agent development, data and tasks, engineering infrastructure, and as a technical lead. \\
\textbf{Federico Carnevale} contributed to agent development and as a sub-effort lead for agent development. \\
\textbf{Petko Georgiev} contributed to agent development, engineering infrastructure, evaluation development, analysis of experiments, writing, and as a technical lead. \\
\textbf{Alex Goldin} contributed to project management. \\
\textbf{Alden Hung} contributed to agent development and as a sub-effort lead for agent development. \\
\textbf{Jessica Landon} contributed to data and tasks, engineering infrastructure, evaluation development, writing, as a sub-effort lead for data and evaluation, and as the STS project lead. \\
\textbf{Timothy Lillicrap} contributed to agent development, data and tasks, environment development, evaluation development, writing, and as an effort lead. \\
\textbf{Alistair Muldal} contributed to data and tasks, evaluation development, analysis of experiments, and as a sub-effort lead for data and evaluation. \\
\textbf{Blake Richards} contributed to writing and analysis of experiments. \\
\textbf{Adam Santoro} contributed to agent development, analysis of experiments, and writing.\\
\textbf{Tamara von Glehn} contributed to agent development and engineering infrastructure. \\
\textbf{Greg Wayne} contributed to agent development, data and tasks, environment development, evaluation development, writing, and as an effort lead. \\
\textbf{Nathaniel Wong} contributed to environment development. \\
\textbf{Chen Yan} contributed to agent development, data and tasks, and engineering infrastructure. \\

\vspace{1mm}
\noindent
{\bf Corresponding Authors:} \\
Jessica Landon (jesslan@deepmind.com) , Greg Wayne (gregwayne@deepmind.com) \& Timothy Lillicrap (countzero@deepmind.com)

\section{Acknowledgments}

The authors would like to thank Tim Harley, Duncan Williams, Arthur Brussee, Mary Cassin, Daan Wierstra, Dario de Cesare, Koray Kavukcuoglu, Matt Botvinick, Lorrayne Bennett, the Worlds Team, and Crowd Compute.

%% file: appendix.tex
\newpage
\section{Appendix}

\subsection{Agent Details}\label{sec:agent_details}
Additional details of the agents highlighted in this work are as follows:

\begin{itemize}
    \item \textbf{MIA:} Behavioral cloning agent with 56M parameters trained on a dataset of human demonstrations in the Playhouse environment. MIA is designed to engage in naturalistic interactions with humans in the Playhouse. For additional details on training and architecture, see ~\cite{iateam2021creating}.
    \item \textbf{MIA, 1/4 data:} MIA ablation using only 1/4 of the amount of data used to train the MIA baseline.
    \item \textbf{MIA, 1/16 data:} MIA ablation using only 1/16 of the amount of data used to train the MIA baseline.
    \item \textbf{MIA, no vision:} MIA ablation that removes visual inputs.
    \item \textbf{large MIA:} MIA ablation with 121M parameters.
    \item \textbf{small MIA:} MIA ablation with 16M parameters.
    \item \textbf{MIA+:} MIA agent with a buffer that compiles language all emissions up to the current step as an input (as opposed to just the most recent language emission).
    \item \textbf{MIA*:} MIA agent with a transformer to pre-process language inputs, an MLP movement policy (instead of LSTM), and previous action as an input.
    \item \textbf{MIA**:} MIA* agent with 165M parameters and dropout.
\end{itemize}

\subsection{Categorization}\label{sec:categorization}

\begin{figure}[ht!]
    \centering
    \includegraphics[width=.9\textwidth]{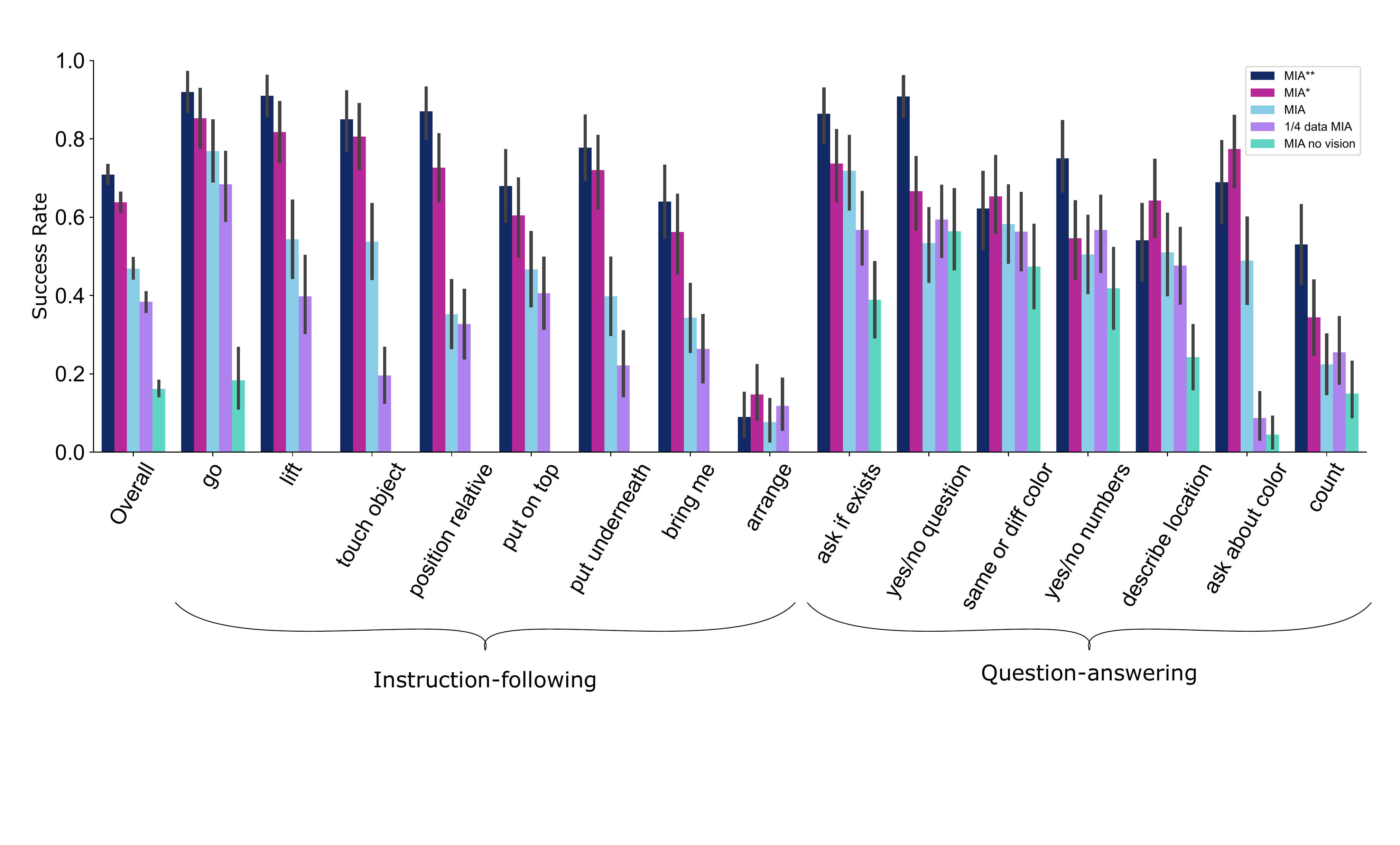}
    \caption{\small{\textbf{Success rate by category}. The \acronym \space enables ranking by both overall score and by any desired sub-category. Here, we show a further breakdown by specific prompt. Viewing these more granular results enables us to identify particular weaknesses in agents. For example, most of our agents are bad at tasks requiring them to ``arrange'' objects, but our best agent is very good at answering ``yes/no questions''.
    }}
    \label{fig:success_by_category}
\end{figure}

\subsection{Sources of Human Errors in the \acronym~Workflow}\label{sec:human_errors}

In the following table, we show a breakdown of the type of human errors identified from attempts at solving the STS scenarios. We reviewed a random sample of 100 episodes that have been annotated as failures. The errors originate from two sources: 1) mistakes made by human players, and 2) mistakes made by human annotators when annotating the episodes. As seen from the table there is no single dominant failure mode; for a variety of reasons attempts at solving the scenarios have failed. 

\begin{table}[ht!]
\begin{center}
\small
\begin{tabular}{cccc} 
Error source & Error type & Number of episodes & Description \\
\hline
\hline
\multirow{4}{*}{Human players} & imprecise & 19 & dexterity issues, not paying attention \\
& out of time & 13 & task is not completed in time \\
& empty & 25 & there is no attempt at solving \\
& control issues & 2 & solver could not control the avatar \\
\hline
\multirow{4}{*}{Human annotators} & annotation mistake & 11 & success marked as failure \\ 
& rigidity & 15 & strict about completion criteria \\ 
& ambiguity & 15 & human answer is ambiguous \\ 
\hline
\end{tabular}
\end{center}
\caption{\label{tab:human-errors}Breakdown of human errors in the STS workflow.}
\end{table}

Errors stemming from mistakes made by human players can be attributed to several different failure modes. First, players can be imprecise -- human players will often have dexterity issues, failing to fully rotate objects, or balance issues, placing objects on top of others without balancing them properly and ensuring the object does not fall off. For some scenarios the deadline is tight and unless the player makes a genuine attempt at solving the task from the beginning of the episode, they may not have time to find the object referred to and perform the requested task with it. 

Errors from human annotators can be due to a genuine mistake or one of two more explainable causes. Scenarios and continuations can sometimes be ambiguous. The primary sources of ambiguity are disagreements about the colors of objects (most commonly shades of pink or off-white) or the number of objects (often some object may not be visible in plain sight unless the player explores the room more thoroughly). Annotators can also be fairly strict about how certain scenarios need to be completed. They will penalize an episode if, for example, the solver enters text in the chat during the interaction unnecessarily. They are also sometimes strict about the positioning of objects when forming shapes and rows. For instance, if the objects are not placed upright (such as potted plants lying sideways rather than placed firmly on the surface), the built shape will be considered invalid.

\subsection{Time to Completion}\label{sec:ttc}

\begin{figure}[ht!]
    \centering
    \includegraphics[width=.6\textwidth]{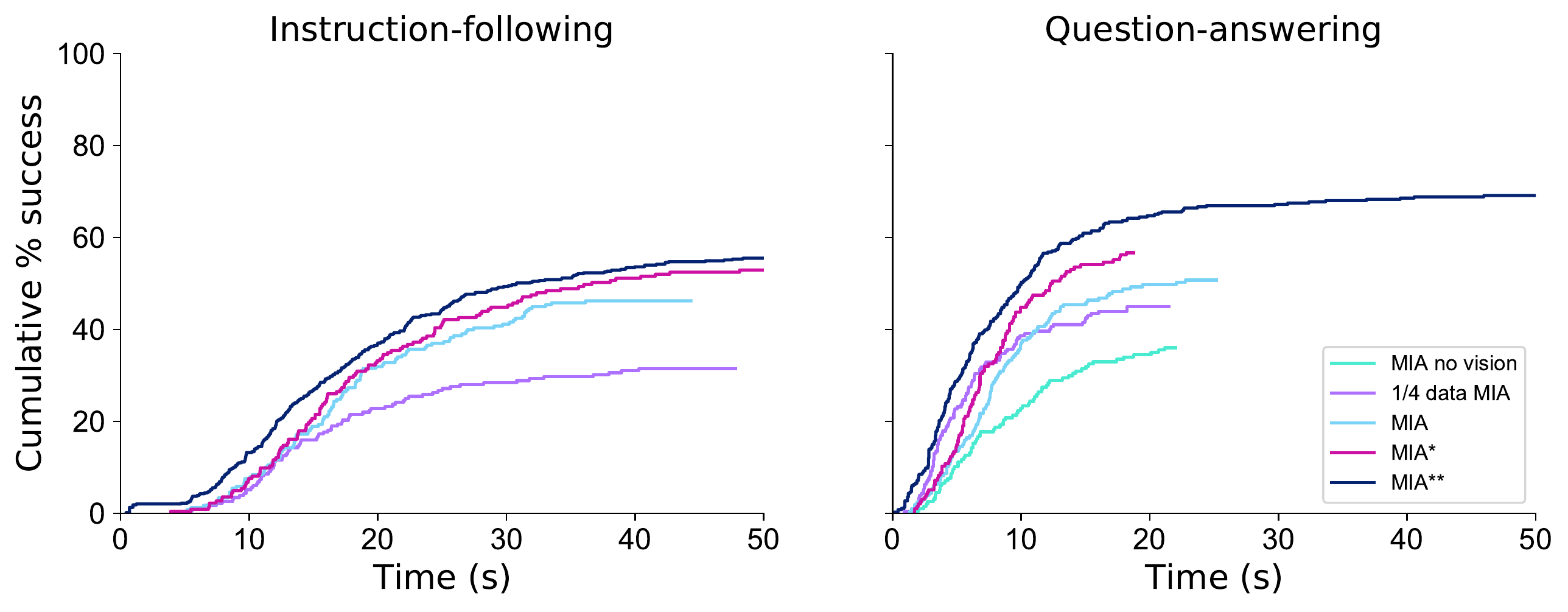}
    \caption{\small{\textbf{CDF of time to completion}. The human annotators annotate the particular time when an agent succeeds in a behavioural continuation. We can use this information to track how quickly agents complete the tasks set across the scenarios in which they succeed.
    }}
    \label{fig:time_to_completion}
\end{figure}

\subsection{Agent Success Rates for an Individual Scenario}\label{sec:consistency}

\begin{figure}[ht!]
    \centering
    \includegraphics[width=.6\textwidth]{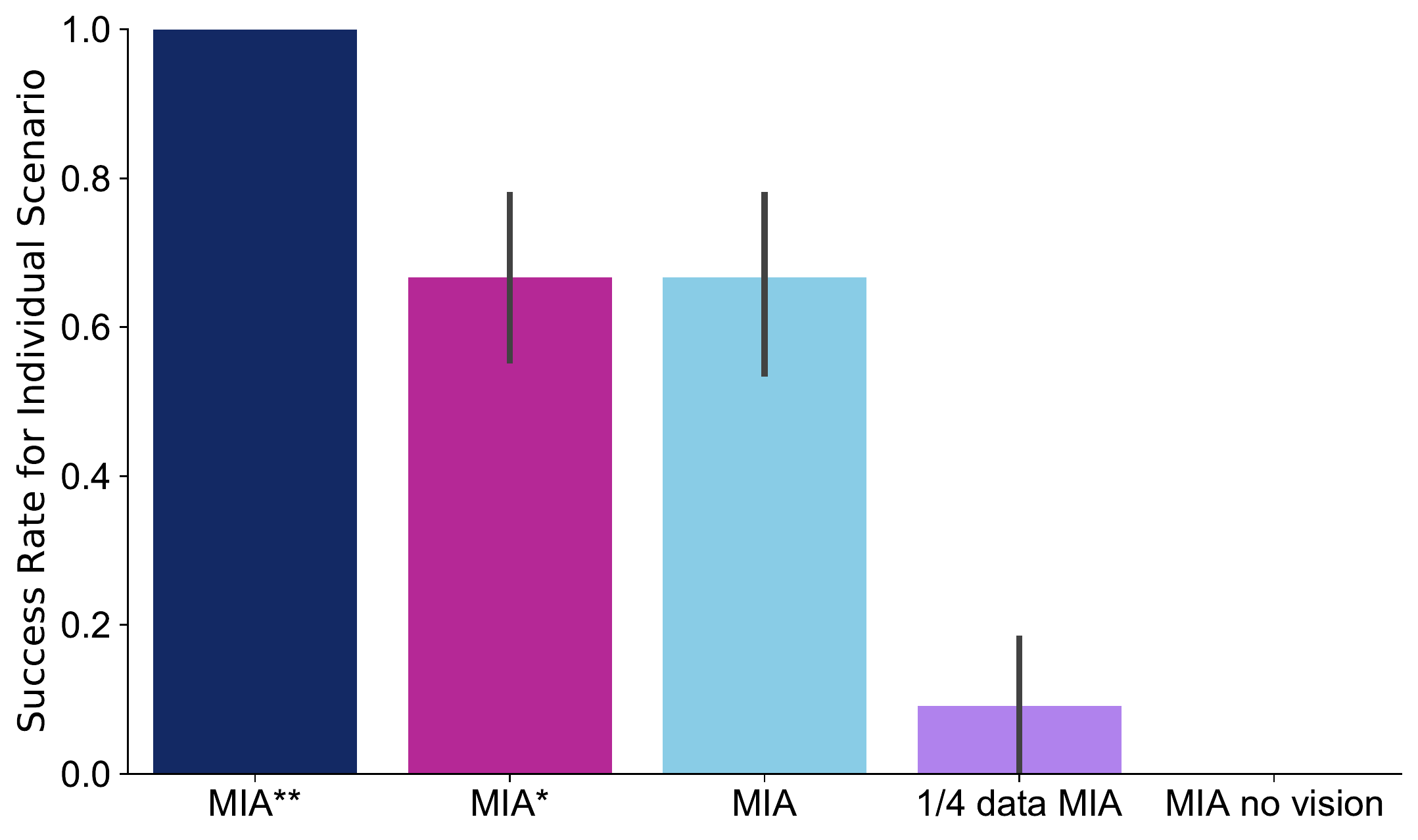}
    \caption{\small{\textbf{Consistency of agents on one scenario}. Because we generate multiple continuations for each individual scenario, we can track how consistently agents succeed. Here, we display the success rate of five agents across ten behavioural continuations on a scenario in the ``lift'' category. Given several agents which are known to have differing competencies, we can use their relative success rates on a particular scenario to measure the difficulty of the scenario. Conversely, we can use an agent's success rate on a given scenario to gauge how competent the agent is compared to other agents.}}
    \label{fig:consistency}
\end{figure}

\subsection{Towards Real-Time Interactions}\label{sec:fps}

\begin{figure}[ht!]
    \centering
    \includegraphics[width=.6\textwidth]{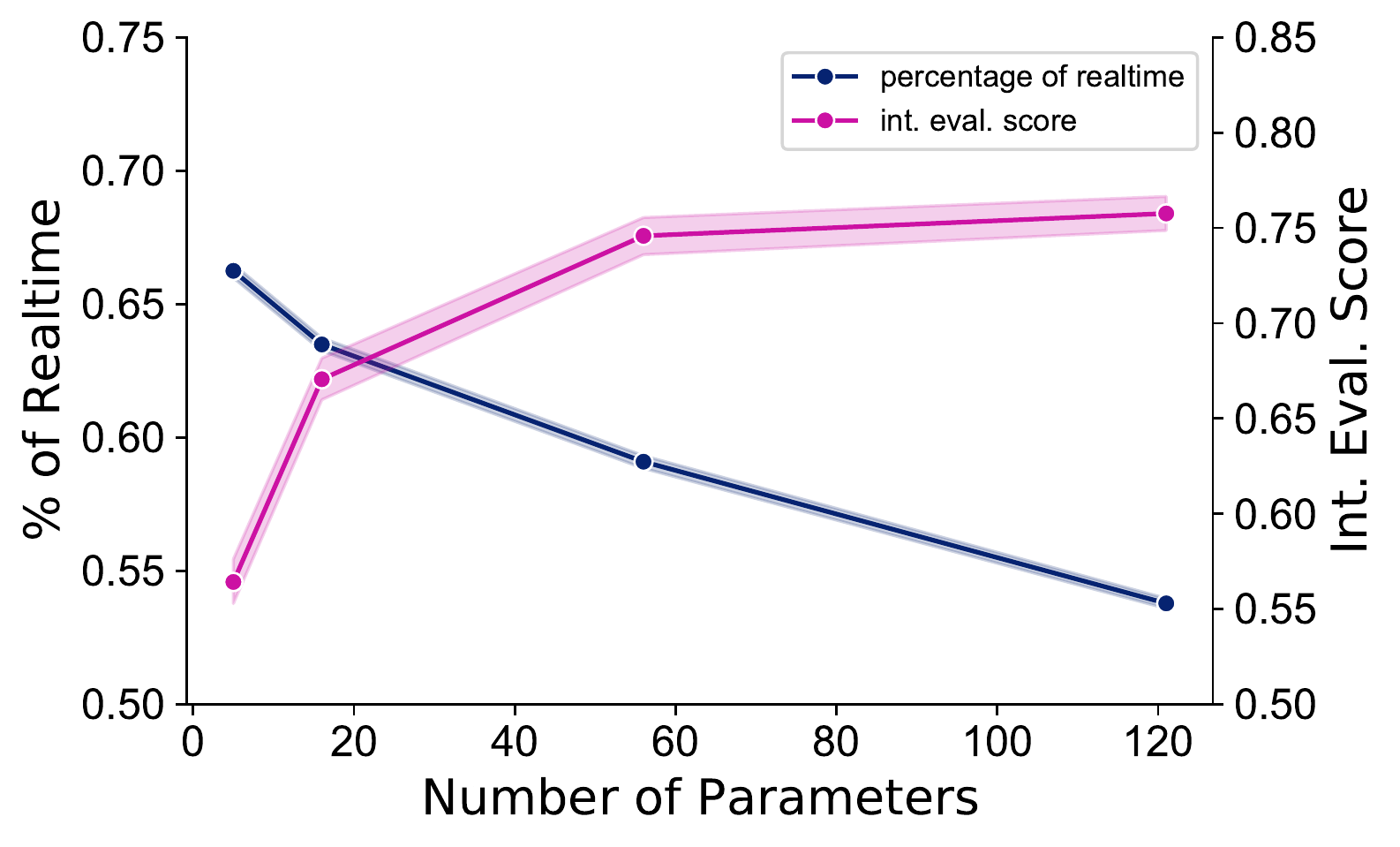}
    \caption{\small{\textbf{Impact of agent size on interactive evaluation score and real-time performance}. Interactive evaluation is predicated on having agents that are fast enough to step in real-time or approximately real-time. As our agents increase in number of parameters, we see that their ability to perform in real-time significantly degrades. Many of our agents run closer to $50\%$ of real-time, and our largest agent, MIA**, is too big to perform in interactive evaluation at all. However, larger agents score better on both interactive evaluation and \acronym~score. Since it is not tied to real-time interactions, the \acronym~allows us to push the boundaries of scale and properly evaluate larger agents. Based on our findings from testing scaling laws, we can then explore ways to make our larger agents faster.
    }}
    \label{fig:fps_scaling}
\end{figure}